\documentclass[sigconf]{acmart}

\usepackage{hyperref}
\usepackage[anythingbreaks]{breakurl}
\def\BibTeX{{\rm B\kern-.05em{\sc i\kern-.025em b}\kern-.08emT\kern-.1667em\lower.7ex\hbox{E}\kern-.125emX}}
    
\copyrightyear{\url{https://xai.kdd2019.a.intuit.com/}}
\acmYear{}
\setcopyright{none}
\acmConference[KDD '19 XAI Workshop]{KDD '19: 25th ACM SIGKDD Conference On Knowledge Discovery and Data Mining}{August 04--08, 2019}{Anchorage, AK,}
\acmBooktitle{KDD '19: 25th ACM SIGKDD Conference On Knowledge Discovery and Data Mining, August 04--08, 2019, Anchorage, AK}
\acmPrice{}
\acmDOI{}
\acmISBN{}

\newcommand{\twopartdef}[4]
{
	\left\{
		\begin{array}{ll}
			#1 & \mbox{if } #2 \\
			#3 & \mbox{if } #4
		\end{array}
	\right.
}

\begin{document}

%
% The "title" command has an optional parameter, allowing the author to define a "short title" to be used in page headers.
\title[Explainable Machine Learning]{On the Art and Science of Explainable Machine Learning: \huge{Techniques, Recommendations, and Responsibilities}}

%
% The "author" command and its associated commands are used to define the authors and their affiliations.
% Of note is the shared affiliation of the first two authors, and the "authornote" and "authornotemark" commands
% used to denote shared contribution to the research.
\author{Patrick Hall}
%\authornote{Both authors contributed equally to this research.}
\email{patrick.hall@h2o.ai}
\affiliation{%
  \institution{H2O.ai}
  \city{Mountain View}
  \country{CA}
}

\renewcommand{\shortauthors}{Hall, Patrick}

%
% The abstract is a short summary of the work to be presented in the article.
\begin{abstract}
This text discusses several popular explanatory methods that go beyond the error measurements and plots traditionally used to assess machine learning models. Some of the explanatory methods are accepted tools of the trade while others are rigorously derived and backed by long-standing theory. The methods, decision tree surrogate models, individual conditional expectation (ICE) plots, local interpretable model-agnostic explanations (LIME), partial dependence plots, and Shapley explanations, vary in terms of scope, fidelity, and suitable application domain. Along with descriptions of these methods, this text presents real-world usage recommendations supported by a use case and public, in-depth software examples for reproducibility.
\end{abstract}


% This command processes the author and affiliation and title information and builds
% the first part of the formatted document.
\maketitle

%-------------------------------------------------------------------------------
\section{Introduction}
%-------------------------------------------------------------------------------

The subject of interpretable machine learning is both multifaceted and evolving. Others have previously defined key terms, put forward general motivations for the broader goal of better interpretability, and advocated for stronger scientific rigor for the burgeoning field \cite{been_kim1}, \cite{gilpin2018explaining}, \cite{guidotti2018survey}, \cite{lipton1}, \cite{weller2017challenges}. Following Doshi-Velez and Kim, this discussion uses ``the ability to explain or to present in understandable terms to a human,'' as the definition of \textit{interpretable}. ``When you can no longer keep asking why,'' will serve as the working definition for a \textit{good explanation} of model mechanisms or predictions \cite{gilpin2018explaining}. These two thoughtful characterizations appear to link explanations and interpretability, and the presented methods help practitioners explain interpretable models and other types of popular supervised machine learning models. Specifically, the discussed techniques facilitate:

\begin{itemize}
	\item Human learning from machine learning.
	\item Human appeal of automated model decisions. 
	\item Regulatory compliance.
	\item White-hat hacking and forensic analysis.
\end{itemize}

\begin{figure}[hb]
	\begin{center}
		\includegraphics[scale=0.1]{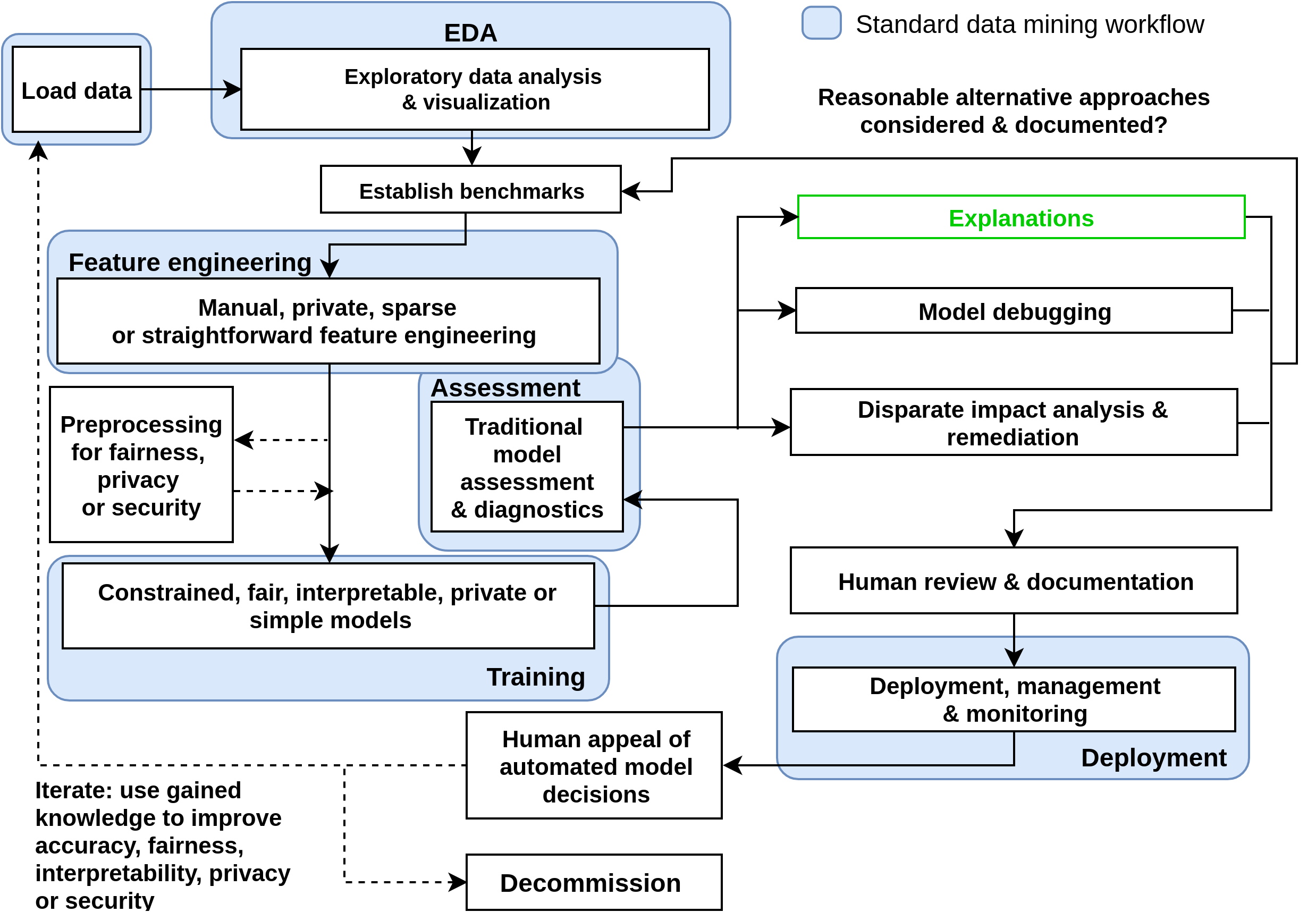}
		\caption{A proposed human-centered workflow in which explanations, and many other techniques, are used to decrease disparate impact and increase interpretability and trust for automated machine learning decision-support systems.}
		\label{fig:hc_ml}
	\end{center}
\end{figure}

Model explanations and the documentation they enable are also an important, mandatory, or embedded aspect of commercial predictive modeling in industries like financial services.\footnote{In the U.S., explanations and model documentation may be required under the Civil Rights Acts of 1964 and 1991, the Americans with Disabilities Act, the Genetic Information Nondiscrimination Act, the Health Insurance Portability and Accountability Act, the Equal Credit Opportunity Act, the Fair Credit Reporting Act, the Fair Housing Act, Federal Reserve SR 11-7, and the European Union (EU) General Data Protection Regulation (GDPR) Article 22 \cite{ff_interpretability}.} However, some have criticized the sub-discipline of explanatory methods like those described herein, and like many technologies, machine learning explanations can be misused, particularly as faulty safeguards for harmful black-boxes and for malevolent purposes like hacking and ``fairwashing'' \cite{fair_washing}, \cite{security_of_ml}, \cite{please_stop}. This text does not condone such practices. It instead promotes informed, nuanced discussion and puts forward best practices to accompany the numerous and already-in-service open source and commercial software packages that have implemented explanatory techniques.\footnote{For a list of open source debugging, explanatory, fairness, and interpretable modeling packages, please see: \url{https://github.com/jphall663/awesome-machine-learning-interpretability}. At this date, commercial implementations include atleast: DataRobot, H2O Driverless AI, SAS Visual Data Mining and Machine Learning, and Zest AutoML.} As highlighted in Figure~\ref{fig:hc_ml}, explanations can, and likely should, be used along with interpretable models, model debugging, disparate impact assessments or bias remediation to further enhance understanding and trust in high-stakes, life- or mission-critical machine learning workflows.

The primary discussion of this text will focus on a pedagogical example scenario that enables detailed description and technical recommendations for the explanatory techniques. A use case will then highlight combining explanations with a constrained, interpretable variant of a gradient boosting machine (GBM). Discussions of the explanatory methods begin below by defining notation. Then Sections ~\ref{sec:surrogate_dt} -- ~\ref{sec:shap} outline explanatory methods and present usage recommendations. Section ~\ref{sec:gen_rec} presents some general interpretability recommendations for practitioners. Section ~\ref{sec:use_case} applies some of the techniques and recommendations to the well-known UCI credit card dataset \cite{uci}. Finally, Section ~\ref{sec:software} highlights software resources that accompany this text. 

%-------------------------------------------------------------------------------
\section{Notation} \label{sec:notation}
%-------------------------------------------------------------------------------

To facilitate descriptions of explanatory techniques, notation for input and output spaces, datasets, and models is defined.

\subsection{Spaces} 
 
	\begin{itemize}
		\item Input features come from the set $\mathcal{X}$ contained in a \textit{P}-dimensional input space, $\mathcal{X} \subset \mathbb{R}^P$.  An arbitrary, potentially unobserved, or future instance of $\mathcal{X}$ is denoted $\mathbf{x}$, $\mathbf{x} \in \mathcal{X}$.
		\item Labels corresponding to instances of $\mathcal{X}$ come from the set $\mathcal{Y}$.
		\item Learned output responses come from the set $\mathcal{\hat{Y}}$. % For regression models, the set $\mathcal{\hat{Y}}_r$ is also contained in a $C$-dimensional output space, $\mathcal{\hat{Y}}_r \subset \mathbb{R}^{C_r}$. For classification models, the set $\mathcal{\hat{Y}}_c$ typically contains a column vector for each unique class in $\mathcal{Y}$. Hence, $\mathcal{\hat{Y}}_c$ is contained in a $C'$-dimensional output space,  $\mathcal{\hat{Y}}_c \subset \mathbb{R}^{C'_c}$.
	\end{itemize}	
	
\subsection{Datasets} 

	\begin{itemize}
		\item The input dataset $\mathbf{X}$ is composed of observed instances of the set $\mathcal{X}$ with a corresponding dataset of labels $\mathbf{Y}$, observed instances of the set $\mathcal{Y}$. 
		\item Each $i$-th observation of $\mathbf{X}$ is denoted as\\ $\mathbf{x}^{(i)} = $  
		$[x_0^{(i)}, x_1^{(i)}, \dots, x_{\textit{P}-1}^{(i)}]$, with corresponding $i$-th labels in $\mathbf{Y}, \mathbf{y}^{(i)}$, and corresponding predictions in $\mathbf{\hat{Y}}, \mathbf{\hat{y}}^{(i)}$. % = [y_0^{(i)}, y_1^{(i)}, \dots, y_{\textit{C}-1}^{(i)}]$.
		\item $\mathbf{X}$ and $\mathbf{Y}$ consist of $N$ tuples of observations:\\ $[(\mathbf{x}^{(0)},\mathbf{y}^{(0)}), (\mathbf{x}^{(1)},\mathbf{y}^{(1)}), \dots,(\mathbf{x}^{(N-1)},\mathbf{y}^{(N-1)})]$. %\\ $\mathbf{x}^{(i)} \in \mathcal{X}$, $\mathbf{y}^{(i)} \in \mathcal{Y}$.
		\item Each $j$-th input column vector of $\mathbf{X}$ is denoted as $X_j = [x_{j}^{(0)}, x_{j}^{(1)}, \dots, x_{j}^{(N-1)}]^T$.
	\end{itemize}	 

\subsection{Models}

	\begin{itemize}
		\item A type of machine learning model $g$, selected from a hypothesis set $\mathcal{H}$, is trained to represent an unknown signal-generating function $f$ observed as  $\mathbf{X}$ with labels $\mathbf{Y}$ using a training algorithm $\mathcal{A}$: 
		$ \mathbf{X}, \mathbf{Y} \xrightarrow{\mathcal{A}} g$, such that $g \approx f$.
		\item $g$ generates learned output responses on the input dataset $g(\mathbf{X}) = \mathbf{\hat{Y}}$, and on the general input space $g(\mathcal{X}) = \mathcal{\hat{Y}}$.
		\item The model to be explained is denoted as $g$.
	\end{itemize}

%-------------------------------------------------------------------------------
\section{Surrogate Decision Trees} \label{sec:surrogate_dt}
%-------------------------------------------------------------------------------

The phrase \textit{surrogate model} is used here to refer to a simple model $h$ of a complex model $g$. This type of model is referred to by various other names, such as \textit{proxy} or \textit{shadow} models and the process of training surrogate models is sometimes referred to as \textit{model extraction} \cite{dt_surrogate2}, \cite{dt_surrogate1}, \cite{ff_interpretability}. 

\subsection{Description}

Given a learned function $g$, a set of learned output responses $g(\mathbf{X}) = \mathbf{\hat{Y}}$, and a tree splitting and pruning approach $\mathcal{A}$, a global -- or over all $\mathbf{X}$ -- surrogate decision tree $h_{\text{tree}}$ can be extracted such that $h_{\text{tree}}(\mathbf{X}) \approx g(\mathbf{X})$:

\begin{equation}
\mathbf{X}, g(\mathbf{X}) \xrightarrow{\mathcal{A}} h_{\text{tree}}
\end{equation}

Decision trees can be represented as directed graphs where the relative positions of input features can provide insight into their importance and interactions \cite{cart}. This makes decision trees useful surrogate models. Input features that appear high and often in the directed graph representation of $h_{\text{tree}}$ are assumed to have high importance in $g$. Input features directly above or below one-another in $h_{\text{tree}}$ are assumed to have potential interactions in $g$. These relative relationships between input features in $h_{\text{tree}}$ can be used to verify and analyze the feature importance, interactions, and predictions of $g$.

\begin{figure}
	\begin{center}
		\includegraphics[scale=0.4]{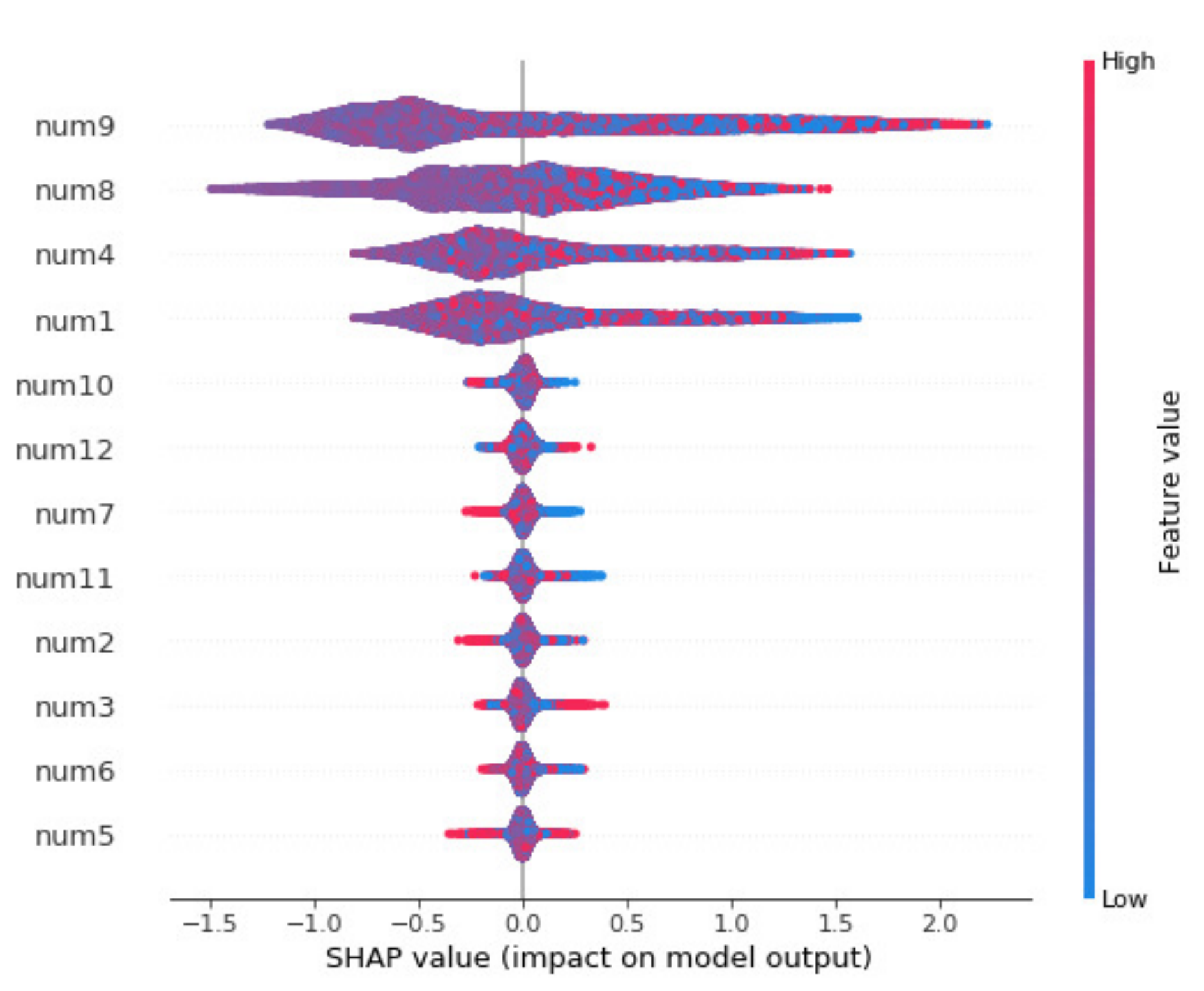}
		\caption{Globally consistent Shapley summary plot for known signal-generating function $f(\mathbf{X}) \sim X_{\text{num}1} * X_{\text{num}4} + |X_{\text{num}8}| * X_{\text{num}9}^2 + e$ and for learned GBM response function $g_{\text{GBM}}$ in a validation dataset.}
		\label{fig:global_shapley}
	\end{center}
\end{figure}

\begin{figure}[htb]
	\begin{center}
		\includegraphics[height=165pt, width=240pt]{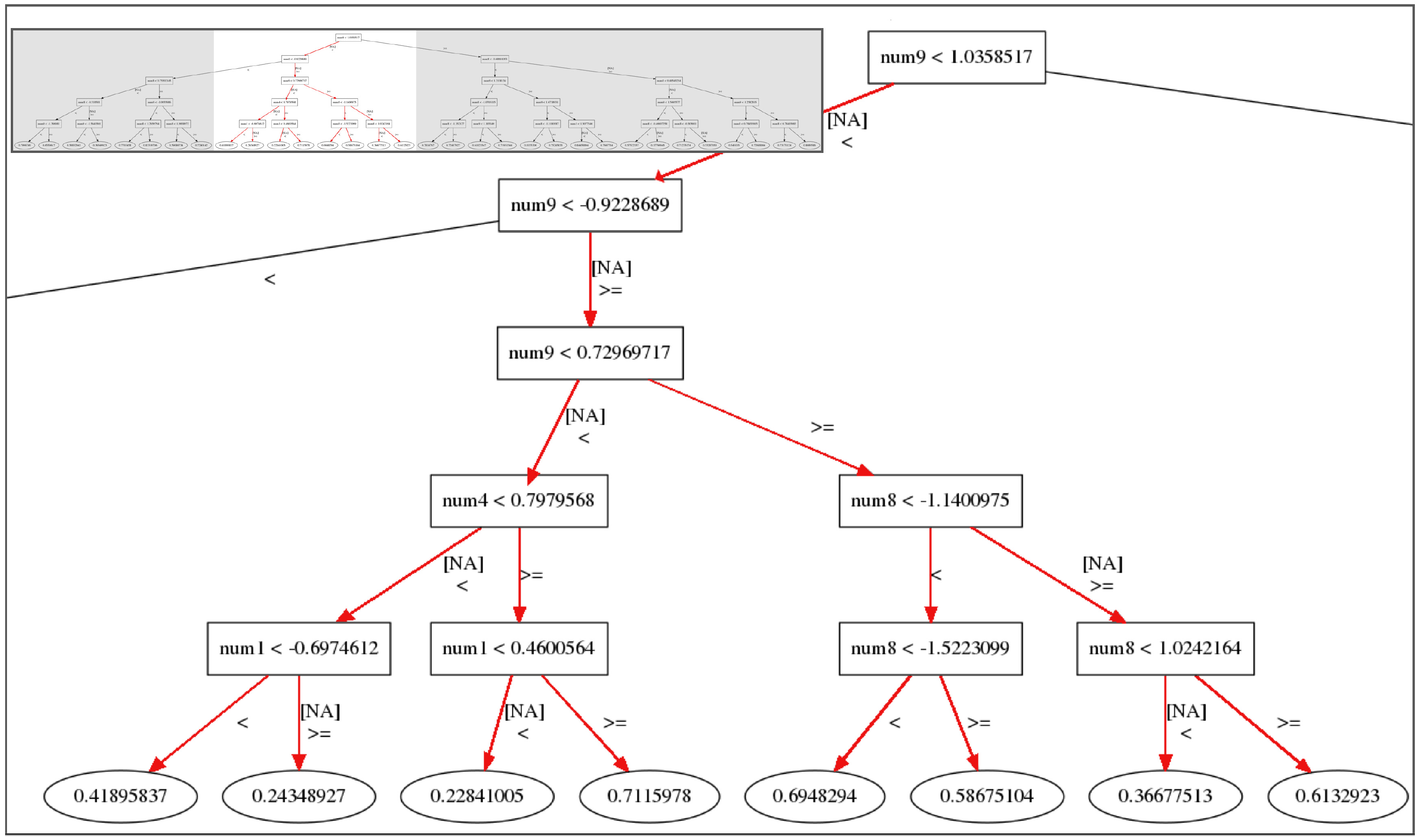}
		\caption{$h_{\text{tree}}$ for previously defined known signal-generating function $f$ and learned GBM response function $g_{\text{GBM}}$ in a validation dataset. An image of the entire $h_{\text{tree}}$ directed graph is available in the supplementary materials described in Section \ref{sec:software}.}
		\label{fig:dt_surrogate}
	\end{center}
\end{figure}

Figures \ref{fig:global_shapley} and \ref{fig:dt_surrogate} use simulated data to empirically demonstrate the desired relationships between input feature importance and interactions in the input space $\mathbf{X}$, a  GBM binary classifier to be explained $g_{\text{GBM}}(\mathbf{X})$, and a decision tree surrogate $h_{\text{tree}}(\mathbf{X})$. Data with a known signal-generating function depending on four numeric input features with interactions and with eight noise features is simulated such that: 

\begin{equation}
\label{eq:f}
f(\mathbf{X}) = \twopartdef {1} {X_{\text{num}1} * X_{\text{num}4} + |X_{\text{num}8}| * X_{\text{num}9}^2 + e \geq 0.42} {0} {X_{\text{num}1} * X_{\text{num}4} + |X_{\text{num}8}| * X_{\text{num}9}^2 + e < 0.42}
\end{equation}

\noindent where $e$ signifies the injection of random noise in the form of label switching for roughly 15\% of the training and validation observations. 

$g_{\text{GBM}}$ is then trained: $ \mathbf{X}, \mathbf{f(X)} \xrightarrow{\mathcal{A}} g_{\text{GBM}}$, such that $g_{\text{GBM}} \approx f$, and $h_{\text{tree}}$ is extracted by: $\mathbf{X}, g_{\text{GBM}}(\mathbf{X}) \xrightarrow{\mathcal{A}} h_{\text{tree}}$,  such that\\ $h_{\text{tree}}(\mathbf{X}) \approx g_{\text{GBM}}(\mathbf{X}) \approx f(\mathbf{X})$.

Figure \ref{fig:global_shapley} displays the local Shapley contribution values, which accurately measure a feature's impact on each $g_{\text{GBM}}(\mathbf{x})$ prediction, for observations in the validation data. Analyzing local Shapley values can be a more holistic and consistent feature importance metric than traditional single-value quantities \cite{shapley}. Features are ordered from top to bottom by their mean absolute Shapley value across observations in Figure \ref{fig:global_shapley}, and as expected, $X_{\text{num}9}$ and $X_{\text{num}8}$ tend to make the largest contributions to $g_{\text{GBM}}(\mathbf{X})$ followed by $X_{\text{num}4}$ and $X_{\text{num}1}$. Also as expected, noise features make minimal contributions to $g_{\text{GBM}}(\mathbf{X})$. Expected values are calculated by training $g_{\text{GBM}}$ with no validation set on $f$ with no error term and are available in materials listed in Section \ref{sec:software}. Shapley values are discussed in detail in Section \ref{sec:shap}. 

Figure \ref{fig:dt_surrogate} is a directed graph representation of $h_{\text{tree}}$ that prominently displays the importance of input features $X_{\text{num}9}$ and $X_{\text{num}8}$ along with $X_{\text{num}4}$ and $X_{\text{num}1}$. Figure \ref{fig:dt_surrogate} also highlights the potential interactions between these inputs. URLs to the data and software used to generate Figures \ref{fig:global_shapley} and \ref{fig:dt_surrogate} are available in Section \ref{sec:software}.

\vspace{-15pt}

\subsection{Recommendations}

\begin{itemize}
	
	\item A shallow-depth $h_{\text{tree}}$ displays a global, low-fidelity (i.e. approximate), high-interpretability flow chart of important features and interactions in $g$. Because there are few theoretical guarantees that $h_{\text{tree}}$ truly represents $g$, always use error measures to assess the trustworthiness of $h_{\text{tree}}$, e.g. RMSE, MAPE, $R^2$.
	
	\item Prescribed methods for training $h_{\text{tree}}$ do exist \cite{dt_surrogate2}, \cite{dt_surrogate1}. In practice, straightforward cross-validation approaches are often sufficient. Moreover, comparing cross-validated training error to traditional training error can give an indication of the stability of the single decision tree $h_{\text{tree}}$.
	
	\item Hu et al. use local linear surrogate models, $h_{\text{GLM}}$, in $h_{\text{tree}}$ leaf nodes to increase overall surrogate model fidelity while also retaining a high degree of interpretability \cite{lime-sup}.
	
\end{itemize}

%-------------------------------------------------------------------------------
\section{Partial Dependence and Individual Conditional Expectation (ICE) Plots}
\label{sec:pd_ice}
%-------------------------------------------------------------------------------

Partial dependence (PD) plots are a widely-used method for describing the average predictions of a complex model $g$ across some partition of data $\mathbf{X}$ for some interesting input feature $X_j$ \cite{esl}. Individual conditional expectation (ICE) plots are a newer method that describes the local behavior of $g$ for a single instance $\mathbf{x} \in \mathcal{X}$. Partial dependence and ICE can be combined in the same plot to compensate for known weaknesses of partial dependence, to identify interactions modeled by $g$, and to create a holistic portrait of the predictions of a complex model for some $X_j$  \cite{ice_plots}.

\subsection{Description}
	
Following Friedman et al. a single feature $X_j \in \mathbf{X}$ and its complement set $\mathbf{X}_{(-j)} \in \mathbf{X}$ (where $X_j \cup \mathbf{X}_{(-j)} = \mathbf{X}$) is considered. $\text{PD}(X_j, g)$ for a given feature $X_j$ is estimated as the average output of the learned function $g(\mathbf{X})$ when all the observations of $X_j$ are set to a constant $x \in \mathcal{X}$ and $\mathbf{X}_{(-j)}$ is left unchanged. $\text{ICE}(x_j, \mathbf{x}, g)$ for a given instance $\mathbf{x}$ and feature $x_j$ is estimated as the output of $g(\mathbf{x})$ when $x_j$ is set to a constant $x \in \mathcal{X}$ and all other features $\mathbf{x} \in \mathbf{X}_{(-j)}$ are left untouched. Partial dependence and ICE curves are usually plotted over some set of constants $x \in \mathcal{X}$. 

\begin{figure}[htb]
	\begin{center}
		\includegraphics[scale=0.45]{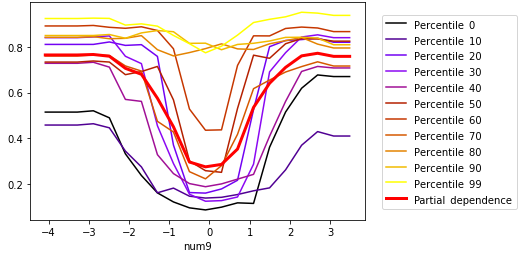}
		\caption{Partial dependence and ICE curves for previously defined known signal-generating function $f$,  learned GBM response function $g_{\text{GBM}}$, and important input feature $X_{\text{num}9}$ in a validation dataset.}
		\label{fig:pdp_ice}
	\end{center}
\end{figure}

As in Section \ref{sec:surrogate_dt}, simulated data is used to highlight desirable characteristics of partial dependence and ICE plots. In Figure \ref{fig:pdp_ice} partial dependence and ICE at the minimum, maximum, and each decile of $g_{\text{GBM}}(\mathbf{X})$ are plotted. The known quadratic behavior of $X_{\text{num}9}$ is plainly visible, except for high value predictions, the 80\textsuperscript{th} percentiles of $g_{\text{GBM}}(\mathbf{X})$ and above and for $\sim-1 < X_{\text{num}9} < \sim1$. When partial dependence and ICE curves diverge, this often points to an interaction that is being averaged out of the partial dependence. Given the form of Equation \ref{eq:f}, there is a known interaction between $X_{\text{num}9}$ and $X_{\text{num}8}$. Combining the information from partial dependence and ICE plots with $h_{tree}$ can help elucidate more detailed information about modeled interactions in $g$. For the simulated example, $h_{tree}$ confirms an interaction between $X_{\text{num}9}$ and $X_{\text{num}8}$ and shows additional modeled interactions between $X_{\text{num}9}$, $X_{\text{num}4}$, and $X_{\text{num}1}$ for $\sim -0.92 \le X_{\text{num}9} <  \sim 1.04.$ URLs to the data and software used to generate Figure \ref{fig:pdp_ice} are available in Section \ref{sec:software}.

\subsection{Recommendations}

\begin{itemize}

\item Combining $h_{\text{tree}}$ with partial dependence and ICE curves is a convenient method for detecting, confirming, and understanding important interactions in $g$.

\item As monotonicity is often a desired trait for interpretable models, partial dependence and ICE plots can be used to verify the monotonicity of $g$ on average and across percentiles of $g(\mathbf{X})$ w.r.t. some input feature $X_j$.

\item Partial dependence can be misleading in the presence of strong correlation or interactions. Plotting partial dependence with ICE provides a direct visual indication as to whether the displayed partial dependence is credibly representative of individual predictions \cite{ice_plots}.

\item Comparing partial dependence and ICE plots with a histogram for the $X_j$ of interest can give a basic qualitative measure of epistemic uncertainty by enabling visual discovery of $g(\mathbf{x})$ values that are based on only small amounts of training data.

\end{itemize}

%-------------------------------------------------------------------------------
\section{Local Interpretable Model-agnostic Explanations (LIME)}
\label{sec:lime}
%-------------------------------------------------------------------------------

Global and local scope are key concepts in explaining machine learning models and predictions. Section \ref{sec:surrogate_dt} presents decision trees as a global -- or over all $\mathbf{X}$ -- surrogate model. As learned response functions, $g$, can be complex, simple global surrogate models can sometimes be too approximate to be trustworthy. LIME attempts to create more representative explanations by fitting a local surrogate model, $h$, in the local region of some observation of interest $\mathbf{x} \in \mathcal{X}$. Both $h$ and local regions can be defined to suit the needs of users.

\subsection{Description}

Ribeiro et al. specifies LIME for some observation $\mathbf{x} \in \mathcal{X}$ as:

\begin{equation}
\begin{aligned}
\underset{h \in \mathcal{H}}{\arg\min}\:\mathcal{L}(g, h, \pi_{\mathbf{X}}) + \Omega(h)
\end{aligned}
\end{equation}

\noindent where $h$ is an interpretable surrogate model of $g$, often a linear model $h_{GLM}$, $\pi_{\mathbf{X}}$ is a weighting function over the domain of $g$, and $\Omega(h)$ limits the complexity of $h$ \cite{lime}. Following Ribeiro et al. $h_{GLM}$ is often trained by:

\begin{equation}
\begin{aligned}
\mathbf{X}', g(\mathbf{X}') \xrightarrow{\mathcal{A}_{\text{LASSO}}} h_{\text{GLM}}
\end{aligned}
\end{equation}

\noindent where $\mathbf{X}'$ is sampled from $\mathcal{X}$, $\pi_{\mathbf{X}}$ weighs $\mathbf{X}'$ samples by their Euclidean similarity to $\mathbf{x}$ to enforce locality, local feature contributions are estimated as the product of $h_{\text{GLM}}$ coefficients and their associated data values $\beta_j x_j$, and $\Omega(h)$ is defined as a LASSO, or L1, penalty on $h_{\text{GLM}}$ coefficients inducing sparsity in $h_{GLM}$. 		

\begin{figure*}[htb]
	\begin{center}
		\includegraphics[scale=0.54]{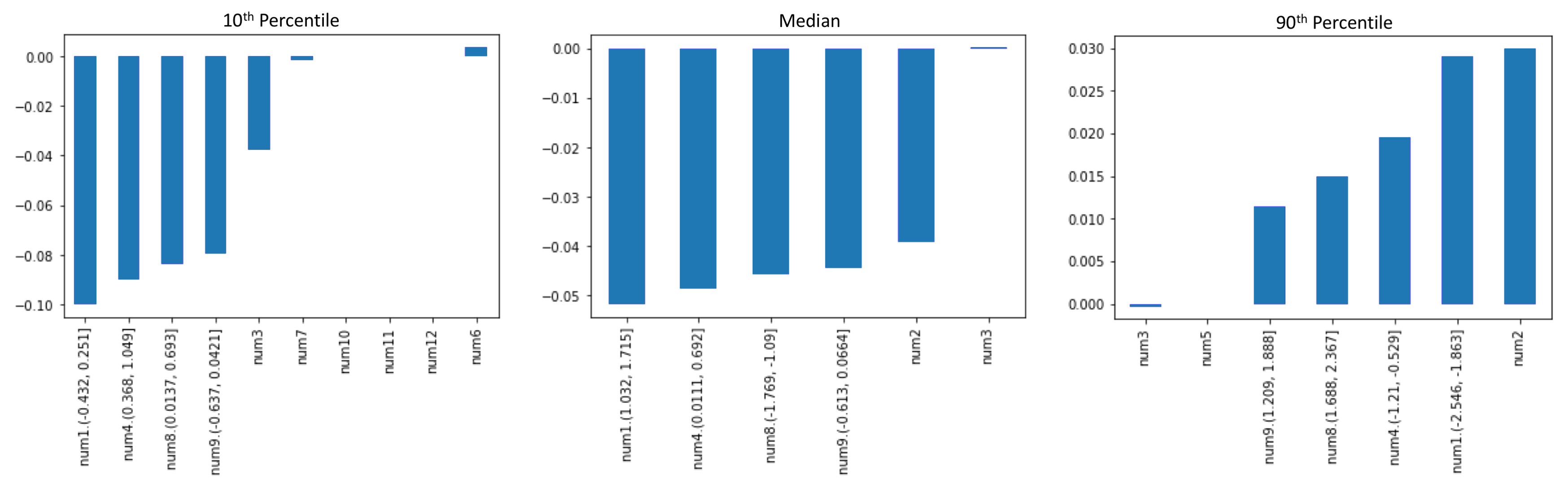}
		\caption{Sparse, low-fidelity local feature contributions found using LIME at three percentiles of $g_{\text{GBM}}(\mathbf{X})$ for known signal-generating function $f(\mathbf{X}) \sim X_{\text{num}1} * X_{\text{num}4} + |X_{\text{num}8}| * X_{\text{num}9}^2 + e$ in a validation dataset.}
		\label{fig:lime}
	\end{center}
\end{figure*}

Figure \ref{fig:lime} displays estimated local feature contribution values for the same $g_{\text{GBM}}$ and simulated $\mathbf{X}$ with known signal-generating function $f$ used in previous sections. To increase the nonlinear capacity of the three $h_{GLM}$ models, information from the Shapley summary plot in Figure \ref{fig:global_shapley} is used to select inputs to discretize before training each $h_{GLM}$: $X_{\text{num}1}, X_{\text{num}4}, X_{\text{num}8}$ and $X_{\text{num}9}$. Table \ref{tab:lime} contains prediction and fit information for $g_{\text{GBM}}$ and $h_{\text{GLM}}$. This is critical information for analyzing LIMEs.

\begin{table}[ht]
	\centering
	\caption{$g_{\text{GBM}}$ and $h_{GLM}$ predictions and $h_{GLM}$ intercepts and $R^2$ for the $h_{GLM}$ models trained to explain $g_{\text{GBM}}(\mathbf{x}^{(i)})$ at the 10\textsuperscript{th}, median, and 90\textsuperscript{th} percentiles of previously defined $g_{\text{GBM}}(\mathbf{X})$ and known signal-generating function $f$ in a validation dataset.} 
	%\tiny
	\begin{tabular}{ | p{1.5cm} | p{1.5cm} | p{1.5cm} | p{1.5cm}| p{0.7cm} | }
	\hline
	$g_{\text{GBM}}(\mathbf{X})$\newline Percentile & $g_{\text{GBM}}(\mathbf{x^{(i)}})$\newline Prediction & $h_{GLM}(\mathbf{x^{(i)}})$\newline Prediction & $h_{GLM}$\newline Intercept & $h_{GLM}$ R\textsuperscript{2} \\ 
	\hline
	10\textsuperscript{th} & 0.16 & 0.13 & 0.53 & 0.72\\
	\hline	
	Median & 0.30 & 0.47 & 0.70 & 0.57\\
	\hline	
	90\textsuperscript{th} & 0.82 & 0.86 & 0.76 & 0.40\\
	\hline
	\end{tabular}
	\label{tab:lime}
\end{table}	

Table \ref{tab:lime} shows that LIME is not necessarily locally accurate, meaning that the predictions of $h_{GLM}(\mathbf{x})$ are not always equal to the prediction of $g_{\text{GBM}}(\mathbf{x})$. Moreover, the three $h_{GLM}$ models do not necessarily explain all of the variance of $g_{\text{GBM}}$ predictions in the local regions around the three $\mathbf{x}^{(i)}$ of interest. $h_{GLM}$ intercepts are also displayed because local feature contribution values, $\beta_j x_j^{(i)}$, are offsets from the local $h_{GLM}$ intercepts.

An immediately noticeable characteristic of the estimated local contributions in Figure \ref{fig:lime} is their sparsity. LASSO input feature selection drives some $h_{GLM}$ $\beta_j$ coefficients to zero so that some $\beta_j x_j^{(i)}$ local feature contributions are also zero. For the 10\textsuperscript{th} percentile $g_{\text{GBM}}(\mathbf{X})$ prediction, the local $h_{GLM}$ R\textsuperscript{2} is adequate and the LIME values appear parsimonious with reasonable expectations. The contributions from discretized $x_{\text{num}1}, x_{\text{num}4}, x_{\text{num}8}$ and $x_{\text{num}9}$ outweigh all other noise feature contributions and the $x_{\text{num}1}, x_{\text{num}4}, x_{\text{num}8}$ and $x_{\text{num}9}$ contributions are all negative as expected for the relatively low value of $g_{\text{GBM}}(\mathbf{x})$. 

For the median prediction of $g_{\text{GBM}}(\mathbf{X})$, it could be expected that some estimated contributions for $x_{\text{num}1}, x_{\text{num}4}, x_{\text{num}8}$ and $x_{\text{num}9}$ should be positive and others should be negative. However, all local feature contributions are negative due to the relatively high value of the $h_{\text{GLM}}$ intercept at the median percentile of $g_{\text{GBM}}(\mathbf{X})$. Because the $h_{\text{GLM}}$ intercept is quite large compared to the $g_{\text{GBM}}(\mathbf{x}^{(i)})$ prediction, it is not alarming that all the $x_{\text{num}1}, x_{\text{num}4}, x_{\text{num}8}$ and $x_{\text{num}9}$ contributions are negative offsets w.r.t. the local $h_{GLM}$ intercept value. For the median $g_{\text{GBM}}(\mathbf{X})$ prediction, $h_{\text{GLM}}$ also estimates that the noise feature $x_{\text{num}2}$ has a fairly large contribution and the local $h_{GLM}$ R\textsuperscript{2} is probably less than adequate to generate fully trustworthy explanations.

For the 90\textsuperscript{th} percentile of $g_{\text{GBM}}(\mathbf{X})$ predictions, the local contributions for $x_{\text{num}1}, x_{\text{num}4}, x_{\text{num}8}$ and $x_{\text{num}9}$ are positive as expected for the relatively high value of $g_{\text{GBM}}(\mathbf{x^{(i)}})$, but the local $h_{GLM}$ R\textsuperscript{2} is somewhat poor and the noise feature $x_{\text{num}2}$ has the highest local feature contribution. This large attribution to the noise feature $x_{\text{num}2}$ could stem from problems in the LIME procedure or in the fit of $g_{\text{GBM}}$ to $f$. Further investigation, or model debugging, is conducted in Section \ref{sec:shap}.

Generally the LIMEs in Section \ref{sec:lime} would be considered to be sparse or high-interpretability but also low-fidelity explanations. This is not always the case with LIME and the fit of some $h_{GLM}$ to a local region around some $g(\mathbf{x})$ will vary in accuracy. URLs to the data and software used to generate Table \ref{tab:lime} and Figure \ref{fig:lime} are available in Section \ref{sec:software}.

\subsection{Recommendations}

\begin{itemize}
	
	\item Always use fit measures to assess the trustworthiness of LIMEs, e.g. RMSE, MAPE, $R^2$.

	\item Local feature contribution values are often offsets from a local $h_{GLM}$ intercept. Note that this intercept can sometimes account for the most important local phenomena. Each LIME feature contribution can be interpreted as the difference in $h(\mathbf{x})$ and some local offset, often $\beta_0$, associated with some feature $x_j$.

	\item Some LIME methods can be difficult to deploy for explaining predictions in real-time. Consider highly deployable variants for real-time applications \cite{h2o_mli_booklet}, \cite{lime-sup}.
		
	\item Always investigate local $h_{GLM}$ intercept values. Generated LIME samples can contain large proportions of out-of-domain data that can lead to unrealistic intercept values. 
		
	\item To increase the fidelity of LIMEs, try LIME on discretized input features and on manually constructed interactions. Analyze $h_{\text{tree}}$ to construct potential interaction terms, and use LIME as further confirmation of modeled $h_{\text{tree}}$ interactions.
 	
	\item Use cross-validation to estimate standard deviations or even confidence intervals for local feature contribution values.
	
	\item Poor fit or inaccuracy of local linear models is itself informative, often indicating extreme nonlinearity or high-degree interactions. However, explanations from such local linear models are likely unacceptable and other types of local models with model-specific explanatory mechanisms, such as decision trees or neural networks, can be used in these cases to generate high-fidelity explanations within the LIME framework \cite{lime}, \cite{wf_xnn}.
	
\end{itemize}

%-------------------------------------------------------------------------------
\section{Tree SHAP} \label{sec:shap}
%-------------------------------------------------------------------------------

Shapley explanations, including tree SHAP (SHapley Additive exPlanations) and even certain implementations of LIME, are a class of additive, locally accurate feature contribution measures with long-standing theoretical support \cite{shapley}. Shapley explanations are the only possible locally accurate and globally consistent feature contribution values, meaning that Shapley explanation values for input features always sum to $g(\mathbf{x})$ and that Shapley explanation values can never decrease for some $x_j$ when $g$ is changed such that $x_j$ truly makes a stronger contribution to $g(\mathbf{x})$ \cite{shapley}. 

\subsection{Description}

For some observation $\mathbf{x} \in \mathcal{X}$, Shapley explanations take the form:

\begin{equation}
\label{eq:shap_additive}
\begin{aligned}
g(\mathbf{x}) = \phi_0 + \sum_{j=0}^{j=\mathcal{P} - 1} \phi_j \mathbf{z}_j
\end{aligned}
\end{equation}

\noindent In Equation \ref{eq:shap_additive}, $\mathbf{z} \in \{0,1\}^\mathcal{P}$ is a binary representation of $\mathbf{x}$ where 0 indicates missingness. Each $\phi_j$ is the local feature contribution value associated with $x_j$ and $\phi_0$ is the average of $g(\mathbf{X})$. 

Shapley values can be estimated in different ways. Tree SHAP is a specific implementation of Shapley explanations. It does not rely on surrogate models. Both tree SHAP and a related technique known as \textit{treeinterpreter} rely instead on traversing internal tree structures to estimate the impact of each $x_j$ for some $g(\mathbf{x})$ of interest \cite{tree_shap}, \cite{treeinterpreter}.

\begin{equation}
\label{eq:shap_contrib}
\begin{aligned}
\phi_{j} = \sum_{S \subseteq \mathcal{P} \setminus \{j\}}\frac{|S|!(\mathcal{P} -|S| -1)!}{\mathcal{P}!}[g_x(S \cup \{j\}) - g_x(S)]
\end{aligned}
\end{equation}

\noindent Unlike treeinterpreter and as displayed in Equation \ref{eq:shap_contrib}, tree SHAP and other Shapley approaches estimate $\phi_j$ as the difference between the model prediction on a subset of features $S$ without $x_j$, $g_x(S)$, and the model prediction with $x_j$ and $S$, $g_x(S \cup \{j\})$, summed and weighed appropriately across all subsets $S$ of $\mathcal{P}$ that do not contain $x_j$, $S \subseteq \mathcal{P} \setminus \{j\}$. (Here $g_x$ incorporates the mapping between $\mathbf{x}$ and the binary vector $\mathbf{z}$.) Since trained decision tree response functions model complex dependencies between input features, removing different subsets of input features helps elucidate the true impact of removing $x_j$ from $g(\mathbf{x})$.

\begin{figure*}[htb]
	\begin{center}
		\includegraphics[scale=0.55]{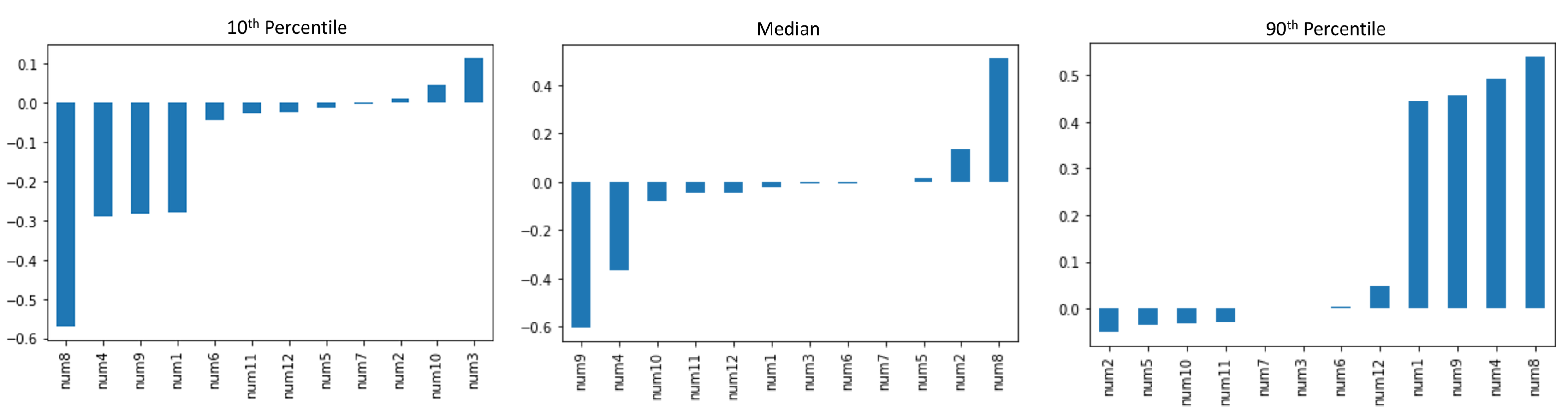}
		\caption{Complete, locally accurate feature contributions found using tree SHAP at three percentiles of $g_{\text{GBM}}(\mathbf{X})$ and for known signal generating function $f(\mathbf{X}) \sim X_{\text{num}1} * X_{\text{num}4} + |X_{\text{num}8}| * X_{\text{num}9}^2 + e$ in a validation dataset.}
		\label{fig:shap}
	\end{center}
\end{figure*}

Simulated data is used again to illustrate the utility of tree SHAP. Shapley explanations are estimated at the 10\textsuperscript{th}, median, and 90\textsuperscript{th} percentiles of $g_{\text{GBM}}(\mathbf{X})$ for simulated $\mathbf{X}$ with known signal-generating function $f$. Results are presented in Figure \ref{fig:shap}. In contrast to the LIME explanations in Figure \ref{fig:lime}, the Shapley explanations are complete, giving a numeric local contribution value for each non-missing input feature. At the 10\textsuperscript{th} percentile of $g_{\text{GBM}}(\mathbf{X})$ predictions, all feature contributions for $x_{\text{num}1}, x_{\text{num}4}, x_{\text{num}8}$ and $x_{\text{num}9}$ are negative as expected for this relatively low value of $g_{\text{GBM}}(\mathbf{X})$ and their contributions obviously outweigh those of the noise features.

For the median prediction of $g_{\text{GBM}}(\mathbf{X})$, the Shapley explanations are somewhat aligned with the expectation of a split between positive and negative contributions. $x_{\text{num}1}, x_{\text{num}4}$, and $x_{\text{num}9}$ are negative and the contribution for $x_{\text{num}8}$ is positive. Like the LIME explanations at this percentile in Figure \ref{fig:lime}, the noise feature $x_{\text{num}2}$ has a relatively high contribution, higher than that of $x_{\text{num}1}$, likely indicating that $g_{\text{GBM}}$ is over-emphasizing $X_{\text{num}2}$ in the local region around the median prediction. 

As expected at the 90\textsuperscript{th} percentile of $g_{\text{GBM}}(\mathbf{X})$ all contributions from $x_{\text{num}1}, x_{\text{num}4}, x_{\text{num}8}$ and $x_{\text{num}9}$ are positive and much larger than the contributions from noise features. Unlike the LIME explanations at the 90\textsuperscript{th} percentile of $g_{\text{GBM}}(\mathbf{X})$ in Figure \ref{fig:lime}, tree SHAP estimates only a small contribution from $x_{\text{num}2}$. This discrepancy may reveal a spurious pair-wise linear correlation between $X_{\text{num}2}$ and $g_{\text{GBM}}(\mathbf{X})$ in the local region around the 90\textsuperscript{th} percentile of $g_{\text{GBM}}(\mathbf{X})$ that fails to represent the true form of $g_{\text{GBM}}(\mathbf{X})$ in this region, which can be highly nonlinear and incorporate high-degree interactions. Partial dependence and ICE for $X_{\text{num}2}$ and two-dimensional partial dependence between $X_{\text{num}2}$ and $X_{\text{num}1}, X_{\text{num}4}, X_{\text{num}8}$ and $X_{\text{num}9}$ could be used to further investigate the form of $g_{\text{GBM}}(\mathbf{X})$ w.r.t. $X_{\text{num}2}$. URLs to the data and software used to generate Figure \ref{fig:shap} are available in Section \ref{sec:software}.

\subsection{Recommendations}

\begin{itemize}
	
	\item Tree SHAP is ideal for estimating high-fidelity, consistent, and complete explanations of decision tree and decision tree ensemble models, perhaps even in regulated applications to generate regulator-mandated reason codes (also known as turn-down codes or adverse action codes).
	
	\item Because tree SHAP explanations are offsets from a global intercept, each $\phi_j$ can be interpreted as the difference in $g(\mathbf{x})$ and the average of $g(\mathbf{X})$ associated with some input feature $x_j$ \cite{molnar}. 
		
	\item Currently treeinterpreter may be inappropriate for some GBM models. Treeinterpreter is locally accurate for some decision tree and random forest models, but is known to be inconsistent like many other feature importance methods aside from Shapley approaches \cite{tree_shap}. In experiments available in the supplemental materials of this text, treeinterpreter is seen to be locally inaccurate for some XGBoost GBM models. 
	
\end{itemize}

%-------------------------------------------------------------------------------
\section{General Recommendations} \label{sec:gen_rec}
%-------------------------------------------------------------------------------

The following recommendations apply to several or all of the described explanatory techniques or to the practice of applied interpretable machine learning in general.

\begin{itemize}	
	
	\item Less complex models are typically easier to explain and several types of machine learning models are directly interpretable, e.g. scalable Bayesian rule lists \cite{sbrl}. For maximum transparency in life- or mission-critical decision support systems, use interpretable white-box machine learning models with model debugging techniques, post-hoc explanatory techniques, and disparate impact analysis and remediation techniques. \\
	
	\item Monotonicity is often a desirable characteristic in interpretable models. (Of course it should not be enforced when a modeled relationship is known to be non-monotonic.) Monotonically constrained \href{https://github.com/dmlc/xgboost}{XGBoost} and \href{https://github.com/h2oai/h2o-3}{h2o-3} GBMs along with the explanatory techniques described in this text are a direct, non-disruptive, open source, and scalable way to train and explain an interpretable machine learning model. A monotonically constrained XGBoost GBM is trained and explained in Section \ref{sec:use_case}.\\
	
	\item Several explanatory techniques are usually required to create good explanations for any given complex model. Users should apply a combination global and local and low- and high-fidelity explanatory techniques to a machine learning model and seek consistent results across multiple explanatory techniques. \\
	
	\item Simpler low-fidelity or sparse explanations can be used to understand more accurate, and sometimes more sophisticated, high-fidelity explanations. \\ 

	\item Methods relying on surrogate models or generated data are sometimes unpalatable to users. Users sometimes \textit{need} to understand \textit{their} model on \textit{their} data.\\
	
	\item Surrogate models can provide low-fidelity explanations for an entire machine learning pipeline in the original feature space if $g$ is defined to include feature extraction or feature engineering steps.\\
	
	\item Understanding and trust are related, but not identical, goals. The discussed explanatory techniques should engender a solid sense of \textit{understanding} in model mechanisms and predictions. Always consider conducting additional model debugging and disparate impact analysis and remediation to foster even greater \textit{trust} in model behavior.\\
	
	\item Consider production deployment of explanatory methods carefully. Currently, the deployment of some open source software packages is not straightforward, especially for the generation of explanations on new data in real-time.\\
	
\end{itemize}

%-------------------------------------------------------------------------------
\section{Credit Card Data Use Case} \label{sec:use_case}
%-------------------------------------------------------------------------------

Some of the discussed explanatory techniques and recommendations will now be applied to a basic credit scoring problem using a monotonically constrained XGBoost binomial classifier and the UCI credit card dataset \cite{uci}. Referring back to Figure \ref{fig:learning_problem}, a training set $\mathbf{X}$ and associated labels $Y$ will be used to train a GBM with decision tree base learners, selected based on domain knowledge from many other types of hypotheses models $\mathcal{H}$, using a monotonic splitting strategy with gradient boosting as the training algorithm $\mathcal{A}_{\text{mono}}$, to learn a final hypothesis model $g_{\text{mono}}$, that approximates the true signal generating function $f$ governing credit default in $\mathcal{X}$ and $\mathcal{Y}$ such that $g_{\text{mono}} \approx f$:

\begin{equation}
\label{eq:cc_training}
\begin{aligned}
\mathbf{X}, \mathbf{Y} \xrightarrow{\mathcal{A_{\text{mono}}}} g_{\text{mono}}
\end{aligned}
\end{equation}

\noindent$g_{\text{mono}}$ is globally explainable with aggregated local Shapley values, decision tree surrogate models $h_{\text{tree}}$, and partial dependence and ICE plots. Additionally each prediction made by $g_{\text{mono}}$ can be explained using local Shapley explanations. 

Thirty percent of the credit card dataset observations are randomly partitioned into a labeled validation set and Pearson correlation between $g_{\text{mono}}$ inputs and the target, $Y_{\text{default payment next month}}$, are calculated and stored. All features except for the target and the observation identifier, $X_{\text{ID}}$, are used as $g_{\text{mono}}$ inputs. The signs of the stored Pearson correlations are used to define the direction of the monotonicity constraints w.r.t. each input feature. (Features with small magnitude correlations or known non-monotonic behavior could also be left unconstrained.) Additional non-default hyperparameter settings used to train $g_{\text{mono}}$ are presented in Table \ref{tab:mono_gbm}. A maximum of 1000 iterations were used to train $g_{\text{mono}}$, with early stopping triggered after 50 iterations without validation AUC improvement. This configuration led to a final validation AUC of 0.781 after only 100 iterations. 

\begin{table}[ht]
	\centering
	\caption{$g_{\text{mono}}$ hyperparameters for the UCI credit card dataset. Adequate hyperparameters were found by Cartesian grid search.}
	%\tiny
	\begin{tabular}{ | p{3.5cm} | p{1.2cm} | }
	\hline
	Hyperparameter & Value \\ 
	\hline
	\texttt{eta} & 0.08 \\
	\hline	
	\texttt{subsample} & 0.9 \\
	\hline	
	\texttt{colsample\_bytree} & 0.9 \\
	\hline
	\texttt{maxdepth} & 15 \\	
	\hline
	\end{tabular}
	\label{tab:mono_gbm}
\end{table}	 

\begin{figure}[htb]
	\begin{center}
		\includegraphics[scale=0.4]{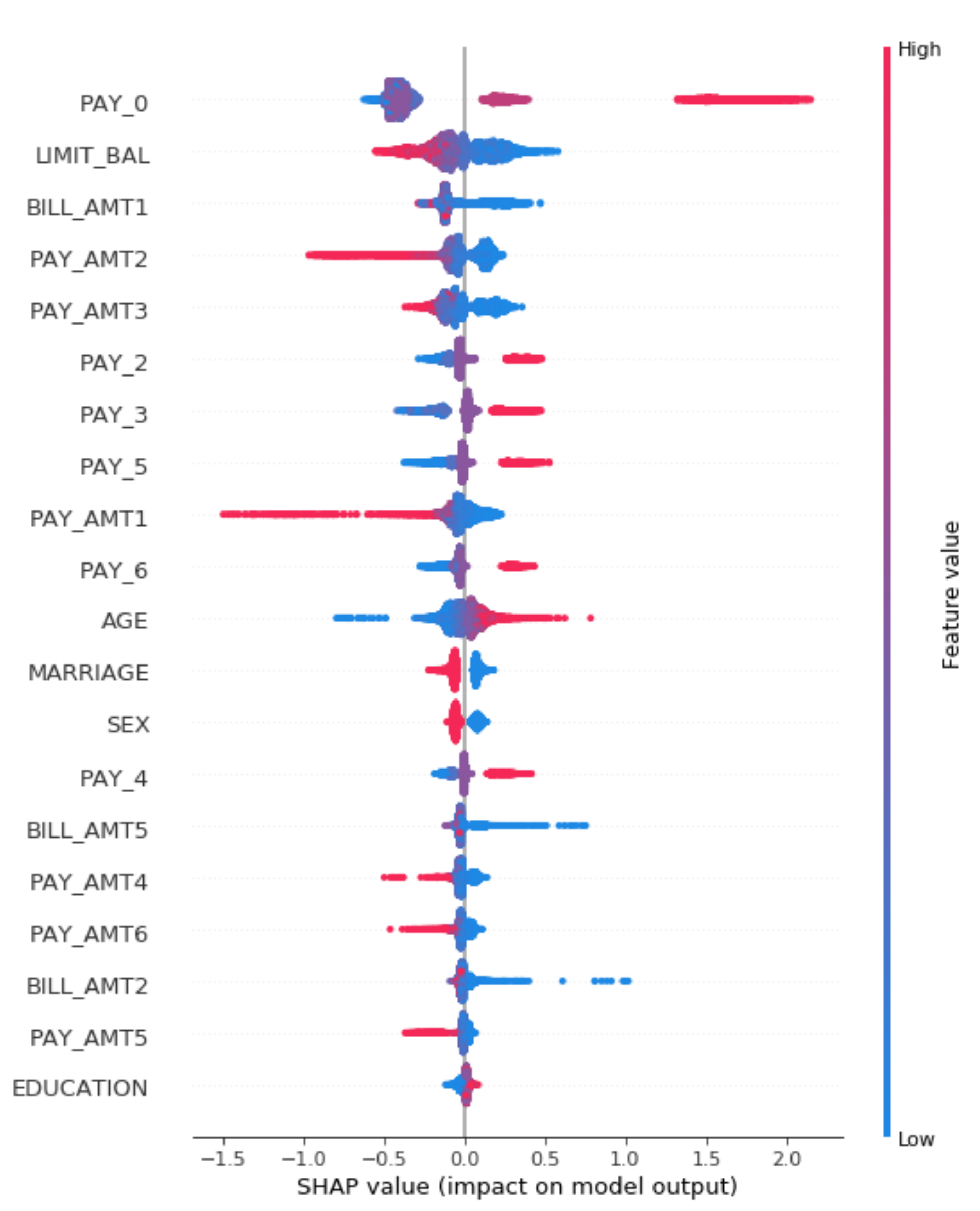}
		\caption{Globally consistent Shapley summary plot for $g_{\text{mono}}$ in a 30\% validation set randomly sampled from the UCI credit card dataset.}
		\label{fig:cc_global_shapley}
	\end{center}
\end{figure}

The global feature importance of $g_{\text{mono}}$ evaluated in the validation set and ranked by mean absolute Shapley value is displayed in Figure \ref{fig:cc_global_shapley}. $X_{\text{PAY\_0}}$ -- a customer's most recent repayment status, $X_{\text{LIMIT\_BAL}}$ -- a customer's credit limit, and $X_{\text{BILL\_AMT1}}$ -- a customer's most recent bill amount are the most important features globally, which aligns with reasonable expectations and basic domain knowledge. (A real-world credit scoring application would be unlikely to use $X_{\text{LIMIT\_BAL}}$ as an input feature because this feature could cause target leakage. $X_{\text{LIMIT\_BAL}}$ is used in this small data example to improve $g_{\text{mono}}$ fit.) The monotonic relationship between each input feature and $g_{\text{mono}}$ output is also visible in Figure \ref{fig:cc_global_shapley}. Numeric Shapley explanation values appear to increase only as an input feature value increases as for $X_{\text{PAY\_0}}$, or vice versa, say for $X_{\text{LIMIT\_BAL}}$. 

\begin{figure}[htb]
	\begin{center}
		\includegraphics[scale=0.45]{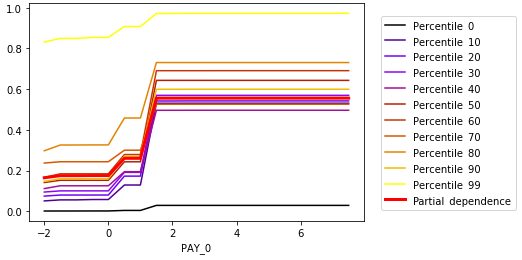}
		\caption{Partial dependence and ICE curves for learned GBM response function $g_{\text{mono}}$ and important input feature $X_{\text{PAY\_0}}$ in a 30\% validation set randomly sampled from the UCI credit card dataset.}
		\label{fig:cc_pdp_ice}
	\end{center}
\end{figure}

Partial dependence and ICE for $g_{\text{mono}}$ and the important input feature $X_{\text{PAY\_0}}$ verify the monotonic increasing behavior of $g_{\text{mono}}$ w.r.t. $X_{\text{PAY\_0}}$. For several percentiles of predicted probabilities and on average, the output of $g_{\text{mono}}$ is low for $X_{\text{PAY\_0}}$ values -2 -- 1 then increases dramatically. $X_{\text{PAY\_0}}$ values of -2 -- 1 are associated with on-time or 1 month late payments. A large increase in predicted probability of default occurs at $X_{\text{PAY\_0}} = 2$ and predicted probabilities plateau after $X_{\text{PAY\_0}} = 2$. The lowest and highest predicted probability customers do not display the same precipitous jump in predicted probability at $X_{\text{PAY\_0}} = 2$. If this dissimilar prediction behavior is related to interactions with other input features, that may be evident in a surrogate decision tree model.

\begin{figure}[htb]
	\begin{center}
		\includegraphics[height=105pt, width=240pt]{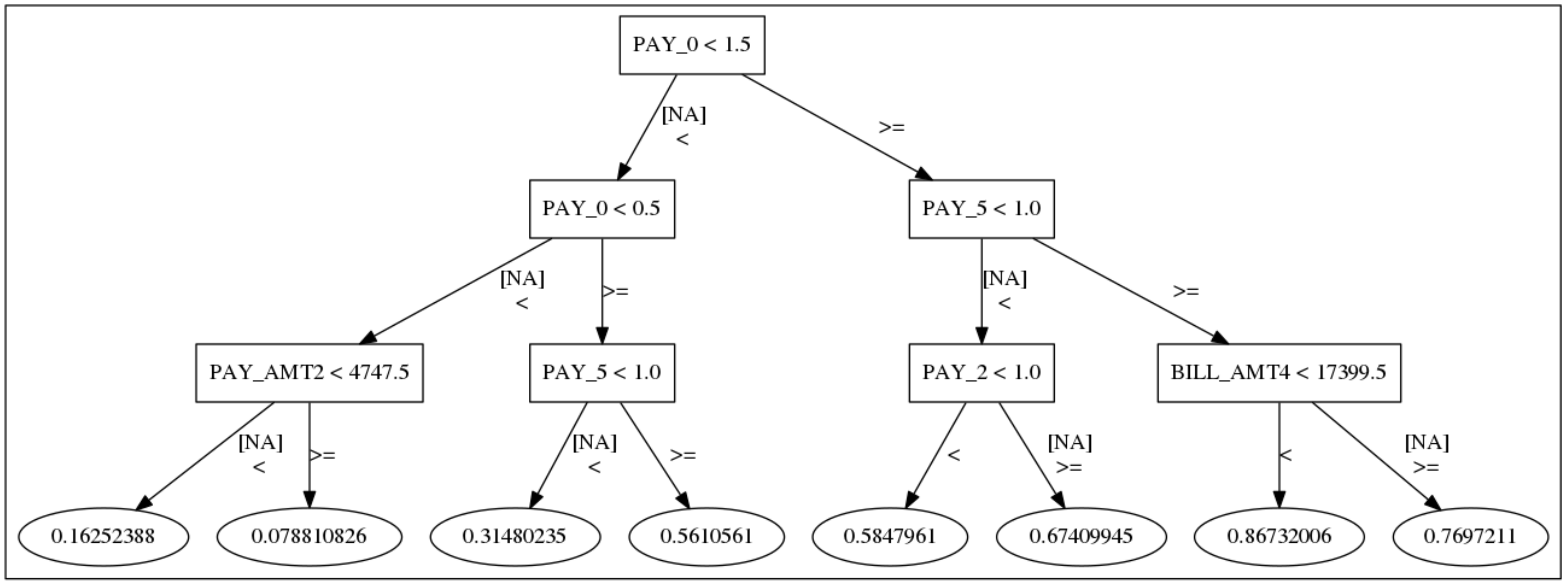}
		\caption{$h_{\text{tree}}$ for $g_{\text{mono}}$ in a 30\% validation set randomly sampled from the UCI credit card dataset. An image of a depth-five $h_{\text{tree}}$ directed graph is available in the supplementary materials described in Section \ref{sec:software}.}
		\label{fig:cc_dt_surrogate}
	\end{center}
\end{figure}

To continue explaining $g_{\text{mono}}$, a simple depth-three $h_{\text{tree}}$ model is trained to represent  $g_{\text{mono}}(\mathbf{X})$ in the validation set. $h_{\text{tree}}$ is displayed in Figure \ref{fig:cc_dt_surrogate}. $h_{\text{tree}}$ has a mean $R^2$ across three random folds in the validation set of 0.86 with a standard deviation of 0.0011 and a mean RMSE across the same folds of 0.08 with a standard deviation of 0.0003, indicating $h_{\text{tree}}$ is likely accurate and stable enough to be a helpful explanatory tool. The global importance of $X_{\text{PAY\_0}}$  and the increase in $g_{\text{mono}}(\mathbf{x})$ associated with $X_{\text{PAY\_0}} = 2$ is reflected in the simple $h_{\text{tree}}$ model, along with several potentially important interactions between input features. For instance the lowest predicted probabilities from $h_{\text{tree}}$ occur when a customer's  most recent repayment status, $X_{\text{PAY\_0}}$, is less than 0.5 and their second most recent payment amount, $X_{\text{PAY\_AMT2}}$, is greater than or equal to NT\$ 4747.5. The highest predicted probabilities from $h_{\text{tree}}$ occur when $X_{\text{PAY\_0}} \geq 1.5$, a customer's fifth most recent repayment status, $X_{\text{PAY\_5}}$, is 1 or more months late, and when a customer's fourth most recent bill amount, $X_{\text{BILL\_AMT4}}$, is less than NT\$ 17399.5. In this simple depth-three $h_{\text{tree}}$ model, it appears that an interaction between $X_{\text{PAY\_0}}$ and $X_{\text{PAY\_AMT2}}$ may be leading to the very low probability of default predictions displayed in Figure \ref{fig:cc_pdp_ice}, while interactions between $X_{\text{PAY\_0}}$, $X_{\text{PAY\_5}}$, and $X_{\text{BILL\_AMT4}}$ are potentially associated with the highest predicted probabilities. A more complex and accurate depth-five $h_{\text{tree}}$ model is available in the supplemental materials described in Section \ref{sec:software} and it presents greater detail regarding the interactions and decision paths that could lead to the modeled behavior for the lowest and highest probability of default customers. 
 
\begin{figure*}[htb]
	\begin{center}
		\includegraphics[scale=0.55]{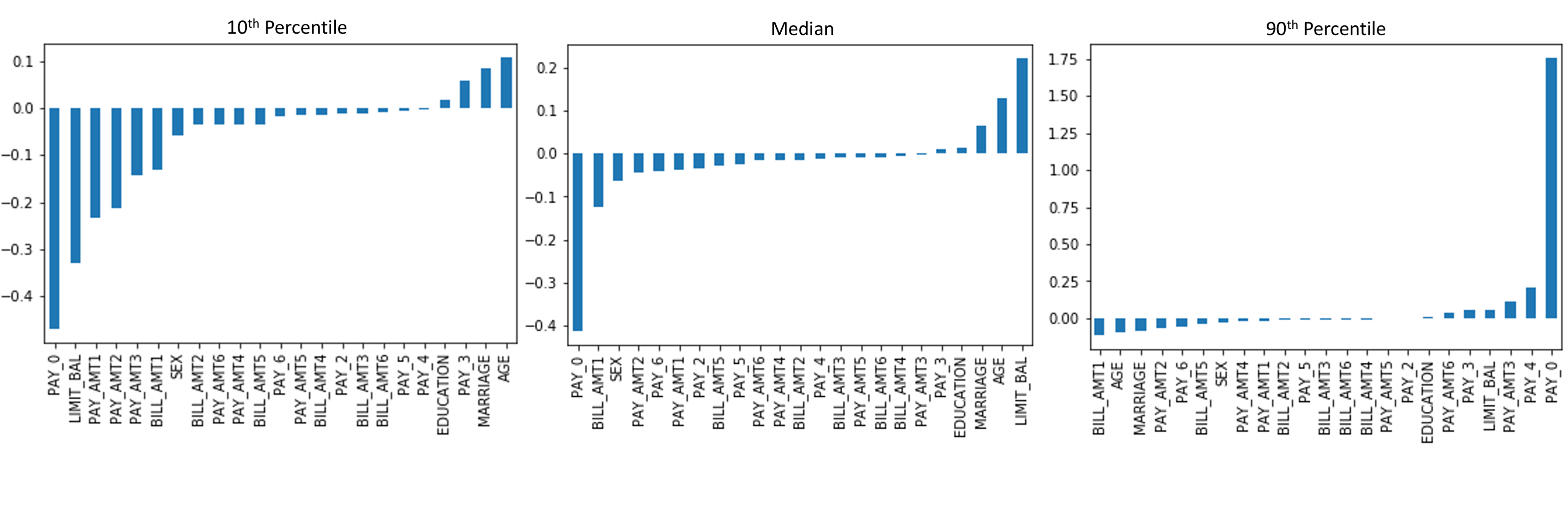}
		\caption{Complete, locally accurate feature contributions found using tree SHAP at three percentiles of $g_{\text{mono}}(\mathbf{X})$ in a 30\% validation set randomly sampled from the UCI credit card dataset.}
		\label{fig:cc_shap}
	\end{center}
\end{figure*}

Figure \ref{fig:cc_shap} displays local Shapley explanation values for three customers at the 10\textsuperscript{th}, median, and 90\textsuperscript{th} percentiles of $g_{\text{mono}}(\mathbf{X})$ in the validation set. The plots in Figure \ref{fig:cc_shap} are representative of the local Shapley explanations that could be generated for any $g_{\text{mono}}(\mathbf{x}), \mathbf{x} \in \mathcal{X}$. The values presented in Figure \ref{fig:cc_shap} are aligned with the general expectation that Shapley contributions will increase for increasing values of $g_{\text{mono}}(\mathbf{x})$. Reason codes to justify decisions based on $g_{\text{mono}}(\mathbf{x})$ predictions can also be generated for arbitrary $g_{\text{mono}}(\mathbf{x})$ using local Shapley explanation values and the values of input features in $\mathbf{x}$. 
Observed values of $\mathbf{x}^{(i)}$ are available in the supplementary materials presented in Section \ref{sec:software}. For the customer at the 90\textsuperscript{th} percentile of $g_{\text{mono}}(\mathbf{X})$ the likely top three reason codes to justify declining further credit are:

\begin{itemize}

\item Most recent payment is 2 months delayed.
\item Fourth most recent payment is 2 months delayed.
\item Third most recent payment amount is NT\$ 0.

\end{itemize} 

Analysis for an operational, mission-critical machine learning model would likely involve further investigation of partial dependence and ICE plots and perhaps deeper analysis of $h_{\text{tree}}$ models following Hu et al \cite{lime-sup}. Analysis would also probably continue on to fairness and model debugging techniques such as:

\begin{itemize}

\item \textbf{Disparate impact analysis and remediation}: to uncover and remediate any disparate impact in model predictions or errors across demographic segments \cite{feldman2015certifying}.
\item \textbf{Residual analysis}: to check the fundamental assumptions of the model against relevant data partitions and investigate outliers or observations exerting undue influence on $g$. 
\item \textbf{Sensitivity analysis}: to explicitly test the trustworthiness of model predictions on simulated out-of-domain data or in other simulated scenarios of interest.  

\end{itemize} 

\noindent A successful explanatory and diagnostic analysis must also include remediating any discovered issues and documenting all findings. Examples of more detailed analyses along with the URLs to the data and software used to generate Figures \ref{fig:cc_global_shapley} -- \ref{fig:cc_shap} are available in Section \ref{sec:software}.

\section{Conclusion}

This text aspires to hasten responsible adoption of explanatory techniques and also to bring interpretable models and model debugging and fairness methodologies to the attention of thoughtful practitioners. Future work will analyze and test combinations of interpretable models and explanatory, model debugging, and fairness techniques in the context of creating accurate and transparent systems for life- or mission-critical decision support applications. 

\section{Acknowledgments}

The author wishes to thank makers past and present at H2O.ai for their tireless support, and especially Sri Ambati, Mark Chan, Doug Deloy, Navdeep Gill, and Wen Phan. The author also thanks Mike Williams of Cloudera Fast Forward Labs for his review of this text.  

% The next two lines define the bibliography style to be used, and the bibliography file.
\bibliographystyle{ACM-Reference-Format}
\bibliography{kdd_2019}

%%% -*-BibTeX-*-
%%% Do NOT edit. File created by BibTeX with style
%%% ACM-Reference-Format-Journals [18-Jan-2012].

\begin{thebibliography}{25}

%%% ====================================================================
%%% NOTE TO THE USER: you can override these defaults by providing
%%% customized versions of any of these macros before the \bibliography
%%% command.  Each of them MUST provide its own final punctuation,
%%% except for \shownote{}, \showDOI{}, and \showURL{}.  The latter two
%%% do not use final punctuation, in order to avoid confusing it with
%%% the Web address.
%%%
%%% To suppress output of a particular field, define its macro to expand
%%% to an empty string, or better, \unskip, like this:
%%%
%%% \newcommand{\showDOI}[1]{\unskip}   % LaTeX syntax
%%%
%%% \def \showDOI #1{\unskip}           % plain TeX syntax
%%%
%%% ====================================================================

\ifx \showCODEN    \undefined \def \showCODEN     #1{\unskip}     \fi
\ifx \showDOI      \undefined \def \showDOI       #1{#1}\fi
\ifx \showISBNx    \undefined \def \showISBNx     #1{\unskip}     \fi
\ifx \showISBNxiii \undefined \def \showISBNxiii  #1{\unskip}     \fi
\ifx \showISSN     \undefined \def \showISSN      #1{\unskip}     \fi
\ifx \showLCCN     \undefined \def \showLCCN      #1{\unskip}     \fi
\ifx \shownote     \undefined \def \shownote      #1{#1}          \fi
\ifx \showarticletitle \undefined \def \showarticletitle #1{#1}   \fi
\ifx \showURL      \undefined \def \showURL       {\relax}        \fi
% The following commands are used for tagged output and should be
% invisible to TeX
\providecommand\bibfield[2]{#2}
\providecommand\bibinfo[2]{#2}
\providecommand\natexlab[1]{#1}
\providecommand\showeprint[2][]{arXiv:#2}

\bibitem[\protect\citeauthoryear{A{\"\i}vodji, Arai, Fortineau, Gambs, Hara,
  and Tapp}{A{\"\i}vodji et~al\mbox{.}}{2019}]%
        {fair_washing}
\bibfield{author}{\bibinfo{person}{Ulrich A{\"\i}vodji},
  \bibinfo{person}{Hiromi Arai}, \bibinfo{person}{Olivier Fortineau},
  \bibinfo{person}{S{\'e}bastien Gambs}, \bibinfo{person}{Satoshi Hara}, {and}
  \bibinfo{person}{Alain Tapp}.} \bibinfo{year}{2019}\natexlab{}.
\newblock \showarticletitle{Fairwashing: the {R}isk of {R}ationalization}.
\newblock \bibinfo{journal}{\emph{arXiv preprint arXiv:1901.09749}}
  (\bibinfo{year}{2019}).
\newblock
\newblock
\shownote{URL: \url{https://arxiv.org/pdf/1901.09749.pdf}.}


\bibitem[\protect\citeauthoryear{Barreno, Nelson, Joseph, and Tygar}{Barreno
  et~al\mbox{.}}{2010}]%
        {security_of_ml}
\bibfield{author}{\bibinfo{person}{Marco Barreno}, \bibinfo{person}{Blaine
  Nelson}, \bibinfo{person}{Anthony~D. Joseph}, {and} \bibinfo{person}{J.~Doug
  Tygar}.} \bibinfo{year}{2010}\natexlab{}.
\newblock \showarticletitle{The {S}ecurity of {M}achine {L}earning}.
\newblock \bibinfo{journal}{\emph{Machine Learning}} \bibinfo{volume}{81},
  \bibinfo{number}{2} (\bibinfo{year}{2010}), \bibinfo{pages}{121--148}.
\newblock
\newblock
\shownote{URL:
  \url{https://people.eecs.berkeley.edu/~adj/publications/paper-files/SecML-MLJ2010.pdf}.}


\bibitem[\protect\citeauthoryear{Bastani, Kim, and Bastani}{Bastani
  et~al\mbox{.}}{2017}]%
        {dt_surrogate2}
\bibfield{author}{\bibinfo{person}{Osbert Bastani}, \bibinfo{person}{Carolyn
  Kim}, {and} \bibinfo{person}{Hamsa Bastani}.}
  \bibinfo{year}{2017}\natexlab{}.
\newblock \showarticletitle{Interpreting {B}lackbox {M}odels via {M}odel
  {E}xtraction}.
\newblock \bibinfo{journal}{\emph{arXiv preprint arXiv:1705.08504}}
  (\bibinfo{year}{2017}).
\newblock
\newblock
\shownote{URL: \url{https://arxiv.org/pdf/1705.08504.pdf}.}


\bibitem[\protect\citeauthoryear{Breiman, Friedman, Olshen, and Stone}{Breiman
  et~al\mbox{.}}{1984}]%
        {cart}
\bibfield{author}{\bibinfo{person}{Leo Breiman}, \bibinfo{person}{Jerome~H.
  Friedman}, \bibinfo{person}{Richard~A. Olshen}, {and}
  \bibinfo{person}{Charles~J. Stone}.} \bibinfo{year}{1984}\natexlab{}.
\newblock \bibinfo{booktitle}{\emph{Classification and Regression Trees}}.
\newblock \bibinfo{publisher}{Routledge}.
\newblock


\bibitem[\protect\citeauthoryear{Craven and Shavlik}{Craven and
  Shavlik}{1996}]%
        {dt_surrogate1}
\bibfield{author}{\bibinfo{person}{Mark~W. Craven} {and}
  \bibinfo{person}{Jude~W. Shavlik}.} \bibinfo{year}{1996}\natexlab{}.
\newblock \showarticletitle{Extracting {T}ree-Structured {R}epresentations of
  {T}rained {N}etworks}.
\newblock \bibinfo{journal}{\emph{Advances in Neural Information Processing
  Systems}} (\bibinfo{year}{1996}).
\newblock
\newblock
\shownote{URL:
  \url{http://papers.nips.cc/paper/1152-extracting-tree-structured-representations-of-trained-networks.pdf}.}


\bibitem[\protect\citeauthoryear{Doshi-Velez and Kim}{Doshi-Velez and
  Kim}{2017}]%
        {been_kim1}
\bibfield{author}{\bibinfo{person}{Finale Doshi-Velez} {and}
  \bibinfo{person}{Been Kim}.} \bibinfo{year}{2017}\natexlab{}.
\newblock \showarticletitle{Towards a {R}igorous {S}cience of {I}nterpretable
  {M}achine {L}earning}.
\newblock \bibinfo{journal}{\emph{arXiv preprint arXiv:1702.08608}}
  (\bibinfo{year}{2017}).
\newblock
\newblock
\shownote{URL: \url{https://arxiv.org/pdf/1702.08608.pdf}.}


\bibitem[\protect\citeauthoryear{Feldman, Friedler, Moeller, Scheidegger, and
  Venkatasubramanian}{Feldman et~al\mbox{.}}{2015}]%
        {feldman2015certifying}
\bibfield{author}{\bibinfo{person}{Michael Feldman},
  \bibinfo{person}{Sorelle~A. Friedler}, \bibinfo{person}{John Moeller},
  \bibinfo{person}{Carlos Scheidegger}, {and} \bibinfo{person}{Suresh
  Venkatasubramanian}.} \bibinfo{year}{2015}\natexlab{}.
\newblock \showarticletitle{Certifying and {R}emoving {D}isparate {I}mpact}. In
  \bibinfo{booktitle}{\emph{Proceedings of the 21th ACM SIGKDD International
  Conference on Knowledge Discovery and Data Mining}}. ACM,
  \bibinfo{pages}{259--268}.
\newblock
\newblock
\shownote{\url{https://arxiv.org/pdf/1412.3756.pdf}.}


\bibitem[\protect\citeauthoryear{Friedman, Hastie, and Tibshirani}{Friedman
  et~al\mbox{.}}{2001}]%
        {esl}
\bibfield{author}{\bibinfo{person}{Jerome Friedman}, \bibinfo{person}{Trevor
  Hastie}, {and} \bibinfo{person}{Robert Tibshirani}.}
  \bibinfo{year}{2001}\natexlab{}.
\newblock \bibinfo{booktitle}{\emph{\textbf{The Elements of Statistical
  Learning}}}.
\newblock \bibinfo{publisher}{Springer}, \bibinfo{address}{New York}.
\newblock
\newblock
\shownote{URL:
  \url{https://web.stanford.edu/~hastie/ElemStatLearn/printings/ESLII\_print12.pdf}.}


\bibitem[\protect\citeauthoryear{Gilpin, Bau, Yuan, Bajwa, Specter, and
  Kagal}{Gilpin et~al\mbox{.}}{2018}]%
        {gilpin2018explaining}
\bibfield{author}{\bibinfo{person}{Leilani~H. Gilpin}, \bibinfo{person}{David
  Bau}, \bibinfo{person}{Ben~Z. Yuan}, \bibinfo{person}{Ayesha Bajwa},
  \bibinfo{person}{Michael Specter}, {and} \bibinfo{person}{Lalana Kagal}.}
  \bibinfo{year}{2018}\natexlab{}.
\newblock \showarticletitle{Explaining {E}xplanations: {A}n {A}pproach to
  {E}valuating {I}nterpretability of {M}achine {L}earning}.
\newblock \bibinfo{journal}{\emph{arXiv preprint arXiv:1806.00069}}
  (\bibinfo{year}{2018}).
\newblock
\newblock
\shownote{URL: \url{https://arxiv.org/pdf/1806.00069.pdf}.}


\bibitem[\protect\citeauthoryear{Goldstein, Kapelner, Bleich, and
  Pitkin}{Goldstein et~al\mbox{.}}{2015}]%
        {ice_plots}
\bibfield{author}{\bibinfo{person}{Alex Goldstein}, \bibinfo{person}{Adam
  Kapelner}, \bibinfo{person}{Justin Bleich}, {and} \bibinfo{person}{Emil
  Pitkin}.} \bibinfo{year}{2015}\natexlab{}.
\newblock \showarticletitle{Peeking {I}nside the {B}lack {B}ox: {V}isualizing
  {S}tatistical {L}earning with {P}lots of {I}ndividual {C}onditional
  {E}xpectation}.
\newblock \bibinfo{journal}{\emph{Journal of Computational and Graphical
  Statistics}} \bibinfo{volume}{24}, \bibinfo{number}{1}
  (\bibinfo{year}{2015}).
\newblock
\newblock
\shownote{URL: \url{https://arxiv.org/pdf/1309.6392.pdf}.}


\bibitem[\protect\citeauthoryear{Guidotti, Monreale, Ruggieri, Turini,
  Giannotti, and Pedreschi}{Guidotti et~al\mbox{.}}{2018}]%
        {guidotti2018survey}
\bibfield{author}{\bibinfo{person}{Riccardo Guidotti}, \bibinfo{person}{Anna
  Monreale}, \bibinfo{person}{Salvatore Ruggieri}, \bibinfo{person}{Franco
  Turini}, \bibinfo{person}{Fosca Giannotti}, {and} \bibinfo{person}{Dino
  Pedreschi}.} \bibinfo{year}{2018}\natexlab{}.
\newblock \showarticletitle{A {S}urvey of {M}ethods for {E}xplaining {B}lack
  {B}ox {M}odels}.
\newblock \bibinfo{journal}{\emph{ACM Computing Surveys (CSUR)}}
  \bibinfo{volume}{51}, \bibinfo{number}{5} (\bibinfo{year}{2018}),
  \bibinfo{pages}{93}.
\newblock
\newblock
\shownote{URL: \url{https://arxiv.org/pdf/1802.01933.pdf}.}


\bibitem[\protect\citeauthoryear{Hall, Gill, Kurka, and Phan}{Hall
  et~al\mbox{.}}{2017}]%
        {h2o_mli_booklet}
\bibfield{author}{\bibinfo{person}{Patrick Hall}, \bibinfo{person}{Navdeep
  Gill}, \bibinfo{person}{Megan Kurka}, {and} \bibinfo{person}{Wen Phan}.}
  \bibinfo{year}{2017}\natexlab{}.
\newblock \bibinfo{booktitle}{\emph{Machine Learning Interpretability with H2O
  Driverless AI}}.
\newblock \bibinfo{publisher}{H2O.ai}.
\newblock
\newblock
\shownote{URL:
  \url{http://docs.h2o.ai/driverless-ai/latest-stable/docs/booklets/MLIBooklet.pdf}.}


\bibitem[\protect\citeauthoryear{Hu, Chen, Nair, and Sudjianto}{Hu
  et~al\mbox{.}}{2018}]%
        {lime-sup}
\bibfield{author}{\bibinfo{person}{Linwei Hu}, \bibinfo{person}{Jie Chen},
  \bibinfo{person}{Vijayan~N. Nair}, {and} \bibinfo{person}{Agus Sudjianto}.}
  \bibinfo{year}{2018}\natexlab{}.
\newblock \showarticletitle{Locally {I}nterpretable {M}odels and {E}ffects
  {B}ased on {S}upervised {P}artitioning ({LIME-SUP})}.
\newblock \bibinfo{journal}{\emph{arXiv preprint arXiv:1806.00663}}
  (\bibinfo{year}{2018}).
\newblock
\newblock
\shownote{URL: \url{https://arxiv.org/ftp/arxiv/papers/1806/1806.00663.pdf}.}


\bibitem[\protect\citeauthoryear{Lichman}{Lichman}{2013}]%
        {uci}
\bibfield{author}{\bibinfo{person}{M. Lichman}.}
  \bibinfo{year}{2013}\natexlab{}.
\newblock \bibinfo{title}{{UCI} {M}achine {L}earning {R}epository}.
\newblock
\newblock
\newblock
\shownote{URL: \url{http://archive.ics.uci.edu/ml}.}


\bibitem[\protect\citeauthoryear{Lipton}{Lipton}{2016}]%
        {lipton1}
\bibfield{author}{\bibinfo{person}{Zachary~C. Lipton}.}
  \bibinfo{year}{2016}\natexlab{}.
\newblock \showarticletitle{The {M}ythos of {M}odel {I}nterpretability}.
\newblock \bibinfo{journal}{\emph{arXiv preprint arXiv:1606.03490}}
  (\bibinfo{year}{2016}).
\newblock
\newblock
\shownote{URL: \url{https://arxiv.org/pdf/1606.03490.pdf}.}


\bibitem[\protect\citeauthoryear{Lundberg, Erion, and Lee}{Lundberg
  et~al\mbox{.}}{2017}]%
        {tree_shap}
\bibfield{author}{\bibinfo{person}{Scott~M. Lundberg},
  \bibinfo{person}{Gabriel~G. Erion}, {and} \bibinfo{person}{Su-In Lee}.}
  \bibinfo{year}{2017}\natexlab{}.
\newblock \showarticletitle{Consistent {I}ndividualized {F}eature {A}ttribution
  for {T}ree {E}nsembles}.
\newblock In \bibinfo{booktitle}{\emph{Proceedings of the 2017 ICML Workshop on
  Human Interpretability in Machine Learning (WHI 2017)}},
  \bibfield{editor}{\bibinfo{person}{Been Kim}, \bibinfo{person}{Dmitry~M.
  Malioutov}, \bibinfo{person}{Kush~R. Varshney}, {and} \bibinfo{person}{Adrian
  Weller}} (Eds.). \bibinfo{publisher}{ICML WHI 2017}, \bibinfo{pages}{15--21}.
\newblock
\newblock
\shownote{URL: \url{https://openreview.net/pdf?id=ByTKSo-m-}.}


\bibitem[\protect\citeauthoryear{Lundberg and Lee}{Lundberg and Lee}{2017}]%
        {shapley}
\bibfield{author}{\bibinfo{person}{Scott~M. Lundberg} {and}
  \bibinfo{person}{Su-In Lee}.} \bibinfo{year}{2017}\natexlab{}.
\newblock \showarticletitle{A {U}nified {A}pproach to {I}nterpreting {M}odel
  {P}redictions}.
\newblock In \bibinfo{booktitle}{\emph{Advances in Neural Information
  Processing Systems 30}}, \bibfield{editor}{\bibinfo{person}{I.~Guyon},
  \bibinfo{person}{U.~V. Luxburg}, \bibinfo{person}{S.~Bengio},
  \bibinfo{person}{H.~Wallach}, \bibinfo{person}{R.~Fergus},
  \bibinfo{person}{S.~Vishwanathan}, {and} \bibinfo{person}{R.~Garnett}}
  (Eds.). \bibinfo{publisher}{Curran Associates, Inc.},
  \bibinfo{pages}{4765--4774}.
\newblock
\newblock
\shownote{URL:
  \url{http://papers.nips.cc/paper/7062-a-unified-approach-to-interpreting-model-predictions.pdf}.}


\bibitem[\protect\citeauthoryear{Molnar}{Molnar}{2018}]%
        {molnar}
\bibfield{author}{\bibinfo{person}{Christoph Molnar}.}
  \bibinfo{year}{2018}\natexlab{}.
\newblock \bibinfo{booktitle}{\emph{Interpretable Machine Learning}}.
\newblock \bibinfo{publisher}{christophm.github.io/interpretable-ml-book}.
\newblock
\newblock
\shownote{URL: \url{https://christophm.github.io/interpretable-ml-book/}.}


\bibitem[\protect\citeauthoryear{Ribeiro, Singh, and Guestrin}{Ribeiro
  et~al\mbox{.}}{2016}]%
        {lime}
\bibfield{author}{\bibinfo{person}{Marco~Tulio Ribeiro},
  \bibinfo{person}{Sameer Singh}, {and} \bibinfo{person}{Carlos Guestrin}.}
  \bibinfo{year}{2016}\natexlab{}.
\newblock \showarticletitle{Why {S}hould {I} {T}rust {Y}ou?: {E}xplaining the
  {P}redictions of {A}ny {C}lassifier}. In
  \bibinfo{booktitle}{\emph{Proceedings of the 22nd ACM SIGKDD International
  Conference on Knowledge Discovery and Data Mining}}. ACM,
  \bibinfo{pages}{1135--1144}.
\newblock
\newblock
\shownote{URL:
  \url{http://www.kdd.org/kdd2016/papers/files/rfp0573-ribeiroA.pdf}.}


\bibitem[\protect\citeauthoryear{Rudin}{Rudin}{2018}]%
        {please_stop}
\bibfield{author}{\bibinfo{person}{Cynthia Rudin}.}
  \bibinfo{year}{2018}\natexlab{}.
\newblock \showarticletitle{Please Stop Explaining Black Box Models for High
  Stakes Decisions}.
\newblock \bibinfo{journal}{\emph{arXiv preprint arXiv:1811.10154}}
  (\bibinfo{year}{2018}).
\newblock
\newblock
\shownote{URL: \url{https://arxiv.org/pdf/1811.10154.pdf}.}


\bibitem[\protect\citeauthoryear{Saabas}{Saabas}{2014}]%
        {treeinterpreter}
\bibfield{author}{\bibinfo{person}{Ando Saabas}.}
  \bibinfo{year}{2014}\natexlab{}.
\newblock \bibinfo{title}{Interpreting {R}andom {F}orests}.
\newblock
\newblock
\newblock
\shownote{URL: \url{http://blog.datadive.net/interpreting-random-forests/}.}


\bibitem[\protect\citeauthoryear{Vaughan, Sudjianto, Brahimi, Chen, and
  Nair}{Vaughan et~al\mbox{.}}{2018}]%
        {wf_xnn}
\bibfield{author}{\bibinfo{person}{Joel Vaughan}, \bibinfo{person}{Agus
  Sudjianto}, \bibinfo{person}{Erind Brahimi}, \bibinfo{person}{Jie Chen},
  {and} \bibinfo{person}{Vijayan~N Nair}.} \bibinfo{year}{2018}\natexlab{}.
\newblock \showarticletitle{Explainable {N}eural {N}etworks {B}ased on
  {A}dditive {I}ndex {M}odels}.
\newblock \bibinfo{journal}{\emph{arXiv preprint arXiv:1806.01933}}
  (\bibinfo{year}{2018}).
\newblock
\newblock
\shownote{URL: \url{https://arxiv.org/pdf/1806.01933.pdf}.}


\bibitem[\protect\citeauthoryear{Weller}{Weller}{2017}]%
        {weller2017challenges}
\bibfield{author}{\bibinfo{person}{Adrian Weller}.}
  \bibinfo{year}{2017}\natexlab{}.
\newblock \showarticletitle{Challenges for {T}ransparency}.
\newblock In \bibinfo{booktitle}{\emph{Proceedings of the 2017 ICML Workshop on
  Human Interpretability in Machine Learning (WHI 2017)}},
  \bibfield{editor}{\bibinfo{person}{Been Kim}, \bibinfo{person}{Dmitry~M.
  Malioutov}, \bibinfo{person}{Kush~R. Varshney}, {and} \bibinfo{person}{Adrian
  Weller}} (Eds.). \bibinfo{publisher}{ICML WHI 2017}, \bibinfo{pages}{55--62}.
\newblock
\newblock
\shownote{URL: \url{https://openreview.net/pdf?id=SJR9L5MQ-}.}


\bibitem[\protect\citeauthoryear{Williams et~al\mbox{.}}{Williams
  et~al\mbox{.}}{2017}]%
        {ff_interpretability}
\bibfield{author}{\bibinfo{person}{Mike Williams} {et~al\mbox{.}}}
  \bibinfo{year}{2017}\natexlab{}.
\newblock \bibinfo{booktitle}{\emph{Interpretability}}.
\newblock \bibinfo{publisher}{Fast Forward Labs}.
\newblock
\newblock
\shownote{URL:
  \url{https://www.cloudera.com/products/fast-forward-labs-research.html}.}


\bibitem[\protect\citeauthoryear{Yang, Rudin, and Seltzer}{Yang
  et~al\mbox{.}}{2017}]%
        {sbrl}
\bibfield{author}{\bibinfo{person}{Hongyu Yang}, \bibinfo{person}{Cynthia
  Rudin}, {and} \bibinfo{person}{Margo Seltzer}.}
  \bibinfo{year}{2017}\natexlab{}.
\newblock \showarticletitle{Scalable {B}ayesian {R}ule {L}ists}. In
  \bibinfo{booktitle}{\emph{Proceedings of the 34th International Conference on
  Machine Learning {(ICML)}}}.
\newblock
\newblock
\shownote{URL: \url{https://arxiv.org/pdf/1602.08610.pdf}.}


\end{thebibliography}

% 
% If your work has an appendix, this is the place to put it.
\appendix
\onecolumn
%-------------------------------------------------------------------------------
\section{Online Software Resources} \label{sec:software}
%-------------------------------------------------------------------------------

To make the discussed results useful and reproducible for practitioners, several online supporting materials and software resources are freely available.\\

\begin{itemize}

	\item Simulated data experiments, including additional experiments on random data, supplementary figures, and the UCI credit card dataset use case are available at:\\
	
		\url{https://github.com/h2oai/mli-resources/tree/master/lime_shap_treeint_compare/README.md}.\\ 
	      
	\item General instructions for using these resources, including a Dockerfile which builds the complete runtime environment with all dependencies, are available here:\\
	      
		\url{https://github.com/h2oai/mli-resources}.\\

	\item In-depth example disparate impact analysis, explanatory, and model debugging use cases for the UCI credit card dataset are available at:\\ 
	
		\url{https://github.com/jphall663/interpretable_machine_learning_with_python}.\\

	\item A curated list of interpretability software is available at:\\
	
		 \url{https://github.com/jphall663/awesome-machine-learning-interpretability}.

\end{itemize}

\end{document}